\documentclass[10pt,twocolumn,letterpaper]{article}

\usepackage{cvpr}              
\usepackage{times}
\usepackage{epsfig}
\usepackage{graphicx}
\usepackage{amsmath}
\usepackage{amssymb}
\usepackage{graphicx}
\usepackage{amsmath}
\usepackage{amssymb}
\usepackage{booktabs}
\usepackage{multirow}
\usepackage{url}
\usepackage{graphicx}
\usepackage{subcaption}

\usepackage{algorithm}
\usepackage{algorithmicx}
\usepackage{algpseudocode} 
\usepackage[dvipsnames]{xcolor}

\definecolor{cvprblue}{rgb}{0.21,0.49,0.74}
\usepackage[pagebackref,breaklinks,colorlinks,citecolor=cvprblue]{hyperref}

\def\paperID{14747} 
\def\confName{CVPR}
\def\confYear{2024}

\title{Task-Adaptive Saliency Guidance for Exemplar-free Class Incremental Learning}
\newcommand{\minisection}[1]{\vspace{0.00in} \noindent {\bf #1}\ }
\author{
Xialei Liu $^{2,1,* }$ \quad
Jiang-Tian Zhai $^{1,}$  \thanks{The first two authors contribute equally.} \quad
Andrew D. Bagdanov $^{3}$ \quad
Ke Li $^{4}$ \quad
Ming-Ming Cheng $^{2,1,}$ \thanks{Corresponding author} \\
\small
$^1$ VCIP, CS, Nankai University \quad
$^2$NKIARI, Shenzhen Futian \quad
$^3$ MICC, University of Florence \quad
$^4$ Tencent Youtu Lab \\
{\tt\small \{xialei,cmm\}@nankai.edu.cn, \{jtzhai30,tristanli.sh\}@gmail.com, andrew.bagdanov@unifi.it }
}

\begin{document}
\maketitle
\begin{abstract}
Exemplar-free Class Incremental Learning (EFCIL) aims to sequentially learn tasks with access only to data from the current one. EFCIL is of interest because it mitigates concerns about privacy and long-term storage of data, while at the same time alleviating the problem of catastrophic forgetting in incremental learning.
In this work, we introduce task-adaptive saliency for EFCIL and propose a new framework, which we call Task-Adaptive Saliency Supervision (TASS), for mitigating the negative effects of saliency drift between different tasks. We first apply boundary-guided saliency to maintain task adaptivity and \textit{plasticity} on model attention. Besides, we introduce task-agnostic low-level signals as auxiliary supervision to increase the \textit{stability} of model attention. Finally, we introduce a module for injecting and recovering saliency noise to increase the robustness of saliency preservation. Our experiments demonstrate that our method can better preserve saliency maps across tasks and achieve state-of-the-art results on the CIFAR-100, Tiny-ImageNet, and ImageNet-Subset EFCIL benchmarks. Code is available at \url{https://github.com/scok30/tass}.
\end{abstract}

\section{Introduction}

Deep neural networks achieve state-of-the-art performance on many computer vision tasks. However, most of these tasks consider a static world in which tasks are well-defined, stationary, and all training data is available in a single training session. The real world consists of dynamically changing environments and data distributions, which -- especially given the computational burden of training large CNNs -- has led to renewed interest in learning new tasks incrementally while avoiding catastrophic forgetting~\cite{goodfellow2013empirical,mccloskey1989catastrophic}.

Class Incremental Learning (CIL)~\cite{belouadah2021comprehensive,masana2020class} is a scenario that considers the possibility of adding new classes to already-trained models. Most CIL methods rely on a memory buffer to store exemplars from past tasks~\cite{castro2018end,douillard2020podnet,rebuffi2017icarl,wu2019large}. In this paper, we consider Exemplar-Free Class Incremental Learning (EFCIL), which is a more challenging setting in which \emph{no data from previous tasks is retained}. This is a realistic scenario and of great interest due to privacy concerns or restrictions on the long-term storage of data. The inability to retain examples from past tasks, however, significantly exacerbates the problem of catastrophic forgetting.

\newcommand{\addle}[1]{\includegraphics[width=0.09\linewidth]{iccv2023AuthorKit/fig/lastsal_withothermethod/#1.png}}
\newcommand{\addfa}[1]{\includegraphics[width=0.11\linewidth]{iccv2023AuthorKit/fig/fig1/#1.png}}
\newcommand{\addfaj}[1]{\includegraphics[width=0.11\linewidth]{iccv2023AuthorKit/fig/fig1/#1.jpg}}
\begin{figure}
	\centering
	\small
	\renewcommand{\arraystretch}{1.1}
  \setlength\tabcolsep{1.2mm}
\small
		\centering
		\includegraphics[width=\columnwidth]{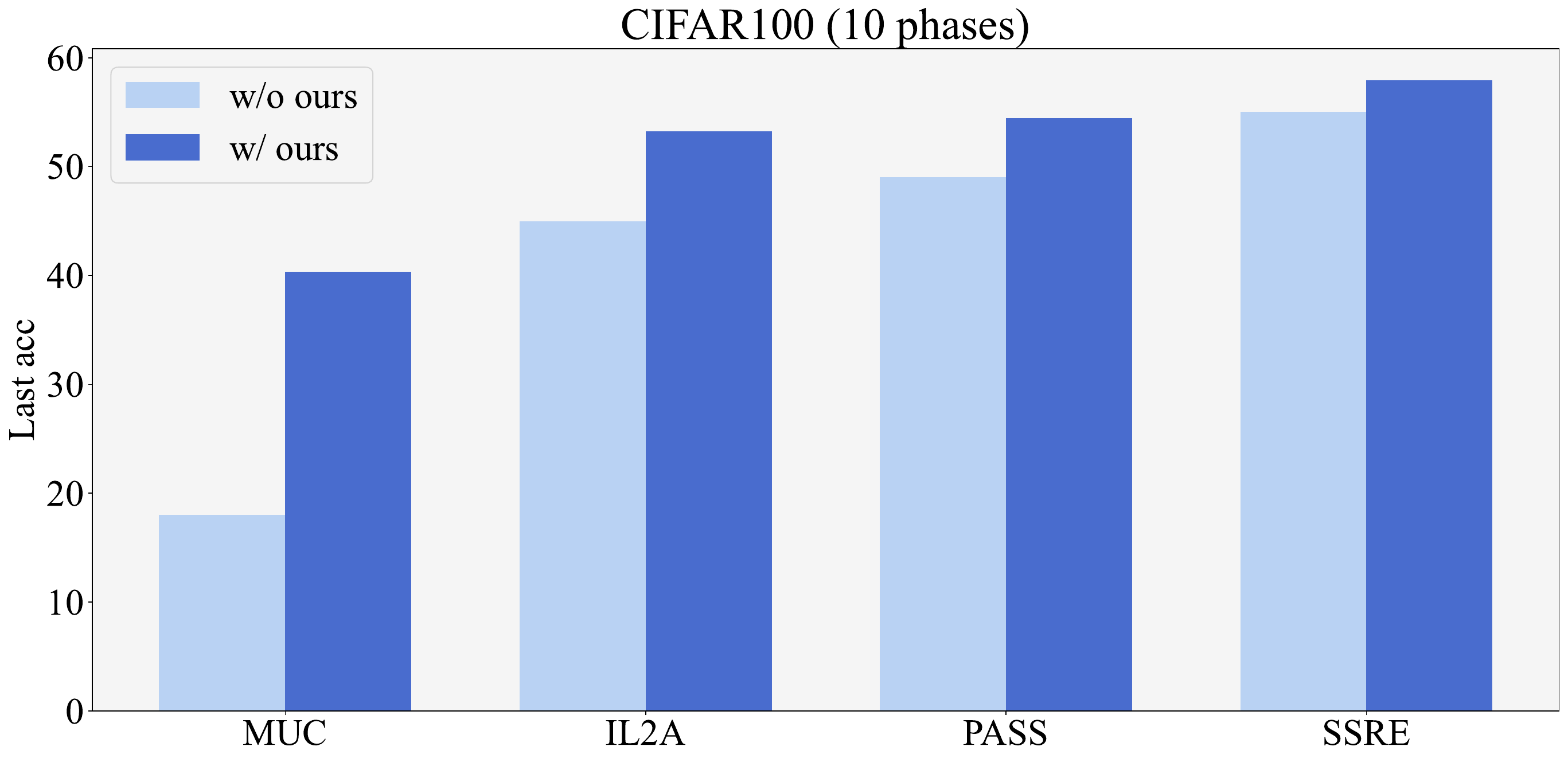}
  \vspace{-20pt}
\caption{We propose the TASS method, which can be directly applied to many recent exemplar-free class incremental learning methods, resulting in a significant improvement in EFCIL classification accuracy and a reduction in catastrophic forgetting.}
\label{fig:fig1}
\vspace{-2mm}
\end{figure}

There are several recent works that consider the EFCIL problem. DeepInversion~\cite{yin2020dreaming} inverts trained networks from random noise to generate images as exemplars and mixes them with current task samples for training. SDC~\cite{yu2020semantic} updates prototypes of each learned class by hypothesizing that semantic drift of classes from previous tasks can be approximated and estimated using new data. Other previous works propose representation learning methods for overcoming catastrophic forgetting~\cite{zhu2021class,zhu2021prototype}. As pointed out in IL2A~\cite{zhu2021class}, learning better representations can reduce representation bias when transferred to new tasks. Incorporating self-supervised learning tasks, such as Barlow Twins~\cite{pham2021dualnet} and rotation prediction~\cite{zhu2021prototype}, has also been proposed to achieve more stable representations and alleviate forgetting. 

CNNs naturally learn to \emph{attend} to features that are discriminative for the tasks they are trained to solve. Catastrophic forgetting also occurs in EFCIL due to the model's attention to \emph{salient} features drifting to features specific to the new task. 
Standard regularization approaches do little to prevent this saliency drift when learning new tasks. One direct method of regularizing saliency is to apply distillation on saliency maps of old samples~\cite{ebrahimi2021remembering}. However, this is complicated by the inability to save samples from previous tasks in the EFCIL setting. Another method is to apply saliency distillation between current task samples and previous task attention~\cite{dhar2019learning}. This method however suffers from the semantic gap between current and old classes when enforcing saliency consistency.

A lack of saliency regularization may lead to attention drifting toward the background in future tasks causing forgetting. Besides, simply applying distillation on attention~\cite{dhar2019learning} fails to offer plasticity and is susceptible to attention forgetting, which is a crucial factor of knowledge forgetting. In comparison,
our Task-Adaptive Saliency Supervision (TASS) approach aims at keeping saliency focused on incrementally learned tasks (for more details, see Section~\ref{sec:experiments}), while maintaining its plasticity and stability. It improves many previous EFCIL methods with higher performance by supervising their attention, as shown in Figure~\ref{fig:fig1}.

Specifically, TASS integrates three components to address this issue. Firstly, we apply dilated boundary maps to prevent saliency drift across object boundaries at intermediate layers. Since saliency drift typically occurs across tasks, encouraging the model to focus on significant foreground regions through dilated boundary supervision reduces the likelihood of saliency shifting toward the background, allowing the model to adaptively select attention areas within the foreground that are relevant to the task. Secondly, to simultaneously enhance the stability of the model's attention across tasks, we add a task-agnostic low-level auxiliary supervision task to the class-incremental framework, which is closely related to our core EFCIL task since image classification has been shown to help models localize the most salient areas within an image. Finally, we propose a module to inject saliency noise into certain feature channels and train the network to denoise them, helping the network further resist attention drift across tasks.

The main contributions of this work are: \textbf{(i)} We provide new insight into task-adaptive saliency supervision under EFCIL settings; we also show the negative effect of methods with no or trivial saliency supervision, which illustrates the superiority of our method and motivates the need for saliency drift mitigation in EFCIL.
\textbf{(ii)} We propose the Task-Adaptive
Saliency Supervision (TASS) with three components that combine to mitigate the saliency drift problem.
\textbf{(iii)} We show that TASS can be easily integrated into other state-of-the-art methods, such as MUC~\cite{liu2020more}, IL2A~\cite{zhu2021class}, PASS~\cite{zhu2021prototype}, SSRE~\cite{zhu2022self}, leading to significant performance gains. 
\textbf{(iv)} Our experiments demonstrate that TASS outperforms all existing EFCIL methods and even several exemplar-based methods on the CIFAR-100, Tiny-ImageNet, and ImageNet-Subset EFCIL benchmarks.

\section{Related Work}
\label{sec:related}

We first discuss previous work on incremental learning from the recent literature and then describe work on EFCIL.

\subsection{Incremental Learning}
A variety of methods have been proposed for incremental learning in the past few years~\cite{delange2021continual,belouadah2021comprehensive}. Recent works can be coarsely grouped into three categories: replay-based, regularization-based, and parameter-isolation methods. Replay-based methods mitigate the task-recency bias by retaining training samples from previous tasks~\cite{rebuffi2017icarl,zhai2023masked}. In addition to replaying samples, BiC~\cite{wu2019large}, PODNet~\cite{douillard2020podnet},  iCaRL~\cite{rebuffi2017icarl} and LTCIL~\cite{liu2022long} apply a distillation loss to prevent forgetting and enhance model stability. GEM~\cite{lopez2017gradient}, AGEM~\cite{AGEM}, and MER~\cite{MER} exploit past-task exemplars by modifying gradients on current training samples to match old samples. Rehearsal may cause models to overfit to stored samples. Regularization-based approaches such as LwF~\cite{li2016learning}, EWC, R-EWC~\cite{kirkpatrick2017overcoming,liu2018rotate}, and DMC~\cite{zhang2020class} offer ways to learn better representations while leaving enough plasticity for adaptation to new tasks. Parameter-isolation methods~\cite{mallya2018packnet,xu2018reinforced} use models with different computational graphs for each task. With the help of growing models, new model branches mitigate catastrophic forgetting at the cost of more parameters and computational cost. It is also widely studied in other fields, such as semantic segmentation~\cite{cermelli2020modeling,lin2023sequential,xiao2023endpoints,yu2022self} and object detection~\cite{feng2022overcoming,liu2023augmented}.

As for saliency-guided incremental learning, LwM~\cite{dhar2019learning} regularizes the saliency activations of previous classes on current task data. However, there exist semantic gaps between current classes and old ones, which results in an inaccurate distillation target for preserving saliency activations on old samples. RRR~\cite{ebrahimi2021remembering} directly saves the Grad-CAM saliency activations of each sample in a replay buffer and applies distillation to memorize this old knowledge, which requires storing additional samples during incremental learning. Despite these initial works on saliency-guided incremental learning, saliency drift still remains problematic and leads to catastrophic forgetting.


\begin{figure*}[t]
		\small
		\centering
  \resizebox{.8\textwidth}{!}{
		\includegraphics[width=\textwidth]{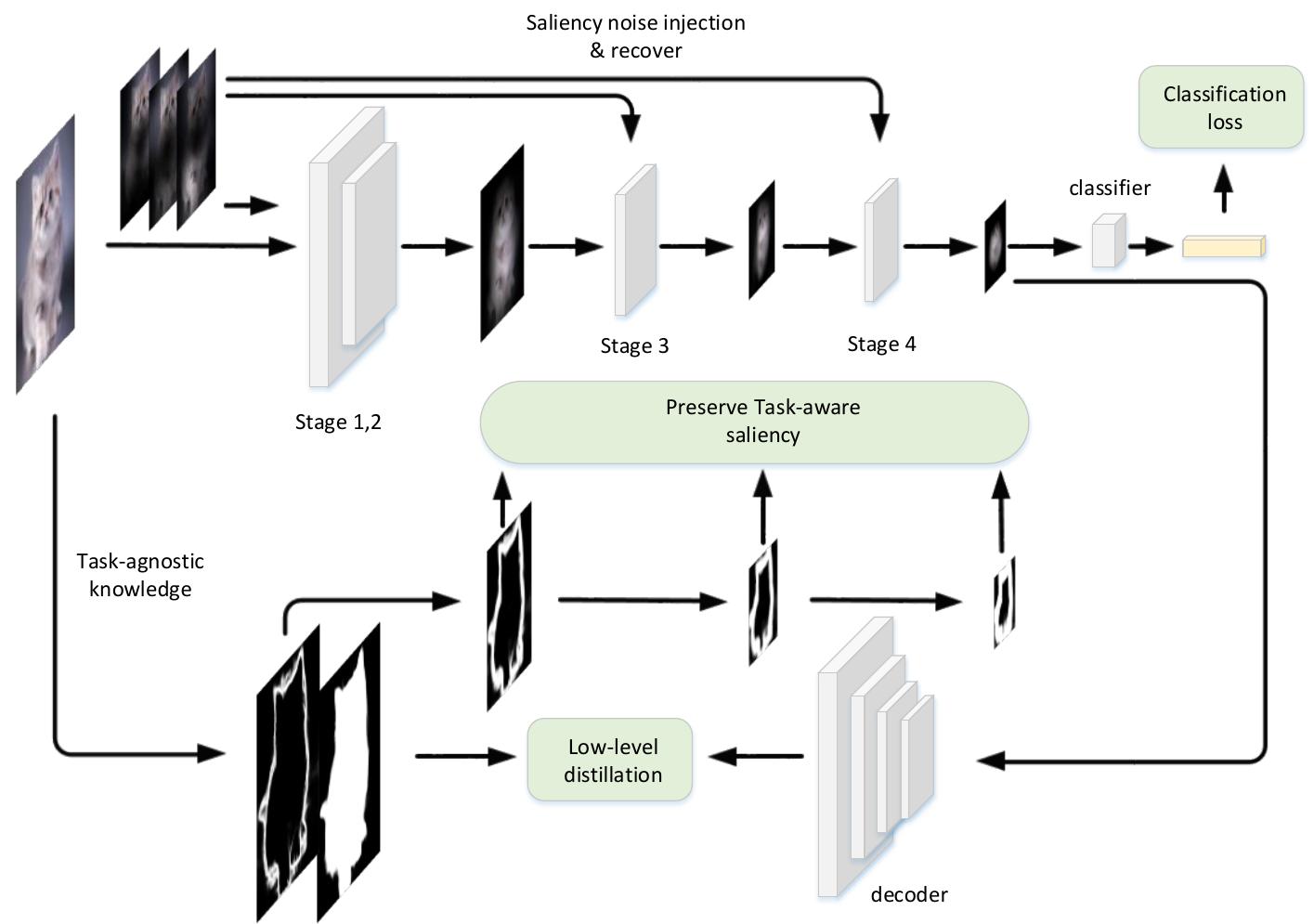}}
\caption{Overall framework of Task-Adaptive Saliency Supervision (TASS). We apply a low-level model to generate saliency and boundary maps. The boundary map is dilated and downsampled to provide supervision at different stages of the encoder. A decoder is attached after the encoder for low-level distillation, which serves as stationary task-agnostic saliency guidance. To prevent saliency drift in later training phases, we introduce saliency noise into each encoder stage. The model is trained to denoise and reduce the saliency drift on current data in future phases. TASS can be integrated into an EFCIL approach to provide robust saliency guidance across incremental tasks.
		}
		\label{Fig:main}
	\end{figure*}
\subsection{Exemplar-free Class Incremental Learning}
Compared to conventional class incremental learning, exemplar-free class incremental learning is more appropriate for applications where training data is sensitive and may not be stored in perpetuity. DAFL~\cite{chen2019data} uses a GAN to generate synthetic samples from past tasks as an alternative to storing actual data. 
DeepInversion, which inverts trained networks using random noise to generate images, is another popular EFCIL method~\cite{yin2020dreaming}. Always Be Dreaming further improves on DeepInversion for EFCIL~\cite{smith2021always}.  
SDC attempts to overcome the problems caused by semantic drift when training new tasks on old class samples~\cite{yu2020semantic}. It directly estimates prototypes of each learned class to use in a nearest class mean classifier. 
PASS~\cite{zhu2021prototype} and IL2A~\cite{zhu2021class} are prototype-based replay methods for efficient and effective EFCIL. Since these prototypes are \textit{features} computed from past training samples, image exemplars need not be retained. SSRE~\cite{zhu2022self} introduces a re-parameterization method to trade-off between old and new knowledge. Our task-Adaptive Saliency Supervision (TASS) method uses three new components to reduce saliency drift in EFCIL and is complementary to several of the approaches mentioned above.



\section{Task-Adaptive Saliency Supervision}
\label{sec:method}

We first define the Exemplar-Free Class Incremental Learning (EFCIL) scenario. Then we describe our TASS approach including dilated boundary supervision, auxiliary low-level supervision, and saliency noise injection. Our overall framework is illustrated in Figure~\ref{Fig:main}.

\subsection{Exemplar-free Class-Incremental Learning}
Class-incremental learning aims to sequentially learn tasks consisting of disjoint classes of samples. Let $t \in \{1,2,..T\}$ denote the incremental learning tasks. The training data $D_t$ for each task contains classes $C_t$ with $N_t$ training samples $\{(x_{t}^{i},y_{t}^{i})\}_{i=1}^{N_t}$, where $x^i_t$ are images and $y_t^i \in C_t$ are their labels.

Most deep networks applied to class-incremental learning can be split into two components: a feature extractor $F_\theta$ and a common classifier $G_\phi$ which grows with each new task $t+1$ to include classes $C_{t+1}$.
The feature extractor $F_\theta$ first maps the input $x$ to a deep feature vector $z=F_\theta(x)\in \mathbb{R}^d$, and then the unified classifier $G_\phi(z) \in \mathbb{R}^{|C_t|}$ is a probability distribution over classes $C_t$ used to make predictions on input $x$.

Class-incremental learning requires that the model be capable of correctly classifying all samples from previous tasks at \textit{any} training task -- that is, when learning task $t$, the model must not forget how to classify samples from classes from tasks $t' < t$. \textit{Exemplar-free} class-incremental learning additionally restricts models to learn each new task without access to samples from previous ones. This typically leads to learning objectives that minimize a loss function $\mathcal{L}$ defined on current training data $D_t$:
\begin{equation}
  \begin{aligned}
    \label{eqn:cil-loss}
    \mathcal{L}_{t}^{\text{CIL}}(x, y) &=   \mathcal{L}_{\text{ce}}(G_{\phi_t}(F_{\theta_t}(x)),y) + \mathcal{L}_{t}^{\text{M}},
  \end{aligned}
\end{equation}
where $\mathcal{L}_{\text{ce}}$ is the standard cross-entropy classification loss and $\mathcal{L}_{t}^{\text{M}}$ is a method-specific loss that mitigates forgetting during incremental learning. Note that \emph{without} $\mathcal{L}_{t}^{\text{M}}$, Eq. \ref{eqn:cil-loss} reduces to fine-tuning on task $t$.

\subsection{Boundary-guided Mid-level Saliency Drift Regularization}

\begin{figure}[t]
    \centering
    \includegraphics[scale=0.26]{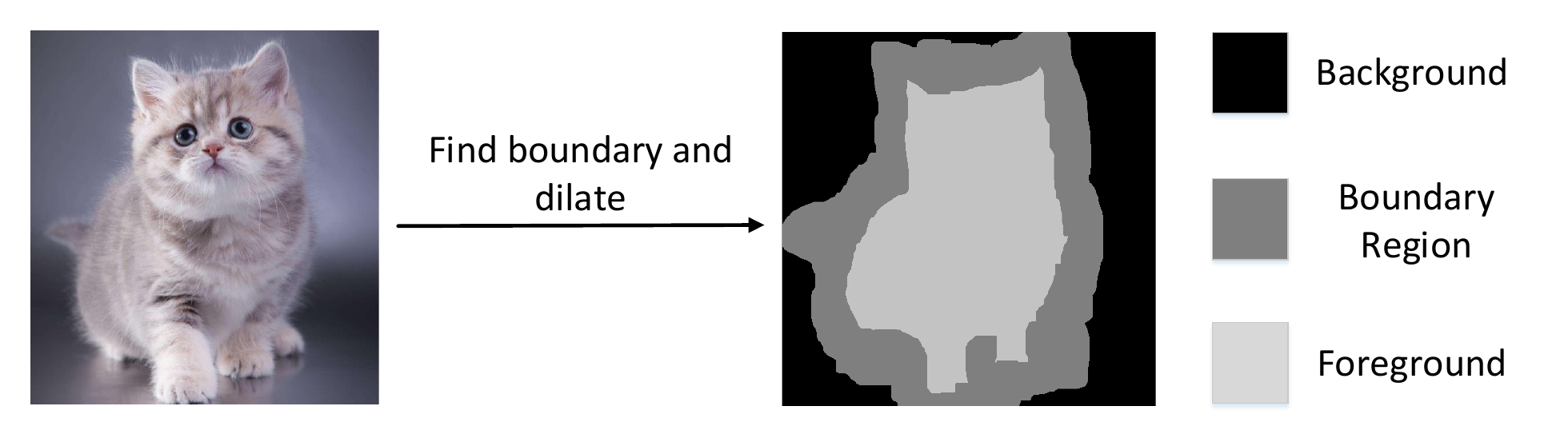}
    \vspace{-20pt}
	\caption{ 
We dilate the boundary map and apply a binary cross entropy loss at three stages in the CNN backbone to prevent mid-level attention from drifting into boundary regions.
	}\label{fig:el}
\end{figure}

Simply distilling attention between tasks does not consider task-adaptive attention.  A model generates low-level representations of each input image $x$ (saliency and boundary maps in our experiments). We use CSNet~\cite{21PAMI-Sal100K} to generate salient regions and boundary maps, as it is lightweight and efficient, but any model producing saliency maps can be used in our framework (we explore other options in the Supplementary Material). To introduce plasticity on attention at these intermediate layers in the backbone, we use the generated boundary maps as a type of adaptive supervision as shown in Figure~\ref{fig:el}. We add a penalty term on object boundaries to avoid drift into the background. We first use 0.5 as the threshold to binarize the generated boundary map and dilate the maps by:
\begin{equation}
  \begin{aligned}
    B_d(x) &= \text{Dilate}(A_b(x),d),
  \end{aligned}
\end{equation}
where $A_b(x)$ is the generated boundary map of image $x$, which is converted from the saliency map with a Laplacian filter, and $d$ denotes the dilation radius applied on the boundary map for controlling the strictness of boundary-guided saliency. 

Rather than use a decoder at each layer as done for the low-level saliency maps described above, the mid-level saliency maps of our model are generated using Grad-CAM~\cite{selvaraju2017grad}\footnote{\scriptsize \url{https://github.com/jacobgil/pytorch-grad-cam}}  at three stages of the CNN backbone (see Figure~\ref{Fig:main}). We also experiment with several other methods for generating student saliency maps and report on them in the Supplementary Material. The dilated generated boundary map $B_d(x)$ is downsampled to match the feature map dimensions at these three stages in order to compare with the Grad-CAM generated saliency boundary maps. We use the binary cross entropy loss for supervision on dilated boundary regions. This loss is defined as:
\begin{equation}
  \begin{aligned}
  \label{eq-dbs}
    \mathcal{L}^{\text{dbs}}_t(x) &= -\frac{\sum_{j=1}^N B_d(x,j) \log(1-S(x,j))}{\sum_{j=1}^N B_d(x,j)},   
  \end{aligned}
\end{equation}
where $S(x,j)$ denotes the saliency map of our model on image $x$ at pixel $j$, $B_d(x,j)$ is the dilated generated boundary map at pixel $j$, and $N$ is the number of pixels in $x$.
We compute this loss only within dilated boundary regions (i.e. where $B_d(x,j)=1$). 
This loss helps the student saliency map avoid intersecting with the dilated teacher boundary region.

\subsection{Auxiliary Low-level Supervisions for EFCIL}
We propose to learn stable features from low-level stationary tasks shared across all incremental classification tasks during class-incremental learning. Low-level vision tasks like salient object detection require useful representations of input images. By learning these feature representations across tasks, the model can focus on key areas of input images and exploit learned, stable features with less representation drift since the low-level features change very little between tasks.

Saliency map prediction is relevant to image classification since the foreground largely determines the results, while the background is comparatively less important. When learning new tasks with new classes, the background of images of new classes may contain new visual concepts that introduce undesirable noise and lead to forgetting of essential previous knowledge. The effectiveness of saliency features for learning classification tasks was demonstrated by Saliency Guided Training~\cite{ismail2021improving}.
Additional supervision of salient region boundaries can aid salient object detection tasks for both segmentation and localization~\cite{fan2023advances,li2023boosting,lin2023learning,ma2024self,zhao2019egnet}. The positive interaction between these two tasks brings richer attention to features relevant to the main classification task. It can provide positive guidance in the form of stationary knowledge across class-incremental tasks. Some examples are illustrated in Figure~\ref{Fig:caf}. 

We incorporate low-level vision tasks into the network as an auxiliary supervision for enriching task-agnostic attention. The boundary map is computed with a Laplacian filter over the estimated saliency map. We add a decoder $D_\psi$~\cite{sfnet} after the backbone $F_\theta$ to predict low-level saliency and boundary maps for input images. The average L2 distance between the prediction and target is used as a low-level saliency distillation loss:
\begin{equation}
\begin{aligned}
\label{eq-lms}
  \mathcal{L}^{\text{lms}}_t(x) &=\frac{||D_{\psi}(F_\theta(x))-A(x)||_2}{\sqrt{N}},
\end{aligned}
\end{equation}
where $A(x)$ denotes the target low-level maps on input $x$, consisting of a saliency map $A_s(x)$ and a boundary map $A_b(x)$. $D_\psi(F_\theta(x))$ are combined saliency and boundary maps produced by the decoder, and $N$ is the number of pixels in the saliency maps.

\subsection{Saliency Noise Injection}

Although we apply low-level distillation and dilated boundary supervision to maintain saliency representations across tasks, the model can still forget saliency on samples from previous tasks. 
To address this, we force the model to be able to recover correct saliency maps from injected saliency noise.

\begin{algorithm}[!t]
	\renewcommand{\algorithmicrequire}{\textbf{Input:}}
	\renewcommand{\algorithmicensure}{\textbf{Output:}}
	\caption{TASS Pseudocode} 
	\label{alg::ross} 
	\begin{algorithmic}[1] 
		\Require 
		The number of tasks $T$, training samples $D_t = \{(x_i, y_i)\}_{i=1}^{N}$ of task $t$, initial parameters $\Theta^0 = \{\theta_0, \phi_0, \psi_0 \}$ containing parameters of feature extractor $F_\theta$, classifier $G_\phi$, and low-level decoder $D_\psi$.
		\Ensure 
		Model $\Theta^T$
		
		\For {$t\in$ $\{1,2,...,T\}$}
		\State $\Theta^t$ ← $\Theta^{t-1}$
		\While {not converged}
        \State Sample $(x, y) \mbox{ from } D_t$
		\State $\mathcal{L}^{\text{CIL}}_t$ ← SaliencyNoiseInjection$(x, y)$
		\State $\mathcal{L}^{\text{lms}}_t$ ← LowLevelMultitask$(x, A(x))$
		\State $S$ ← ComputeGradCAMSaliency($x, y$)
		\State $\mathcal{L}^{\text{dbs}}_t$ ← DilatedBoundarySupervision$(S, A(x))$   
		\State update $\Theta^t$ by minimizing $\mathcal{L}^{\mathrm{all}}_t$ from Eq.~\ref{allloss}
		\EndWhile
		
		\EndFor

	\end{algorithmic} 
\end{algorithm}

At each task there is no available training data from previous or future tasks, and therefore we cannot directly know saliency drift on these samples. Instead of supervising the model with ground-truth saliency drift signals, we introduce saliency noise on random feature channels. We use a random ellipse to approximate the potential saliency drift in future tasks and the model is trained to denoise within each stage. Therefore the model can learn to effectively reduce real saliency drift.

We generate elliptical noise using a very simple approach. There are six parameter dimensions: the center coordinate $(x,y)$, the major and minor axis lengths $(a,b)$, the rotation angle $\alpha$, and the mask weight $w$. A detailed explanation of this process is given in the Supplementary Material. With the help of dilated boundary supervision, each stage learns to eliminate this additional saliency noise and this aids generalization for future tasks and mitigates saliency forgetting in previous ones.


\subsection{Learning Objective and Training Algorithm}

The overall learning objective combines the low-level multi-task learning, dilated boundary supervision, and random saliency noise injection modules:
\begin{equation}
  \begin{aligned}
  \label{allloss}
    \mathcal{L}^{\text{all}}_t &= \mathcal{L}^{\mathrm{CIL}}_t + \mathcal{L}^{\text{lms}}_t +  \mathcal{L}^{\text{dbs}}_t.
  \end{aligned}
\end{equation}
Comparing this loss with Eq.~\ref{eqn:cil-loss}, for TASS $\mathcal{L}_{t}^{\text{M}} = \mathcal{L}^{\mathrm{CIL}}_t + \mathcal{L}^{\text{lms}}_t$, thus incorporating saliency-aware supervision with the cross-entropy loss. The entire process is detailed in Algorithm~\ref{alg::ross}.
		
		 

\begin{table*}[t]
	\centering
	\small
	\renewcommand{\arraystretch}{1.1}
  \setlength\tabcolsep{1.2mm}
\resizebox{.78\textwidth}{!}{	
\begin{tabular}{c|c|c|c|c|c|c}\hline
\multicolumn{1}{r|}{\textbf{Dataset}}&\multicolumn{3}{c|}{\textbf{CIFAR-100}} &\multicolumn{3}{c}{\textbf{Tiny-ImageNet}} \\ \hline\hline
	   \textbf{Method} & \textbf{5 tasks} &  \textbf{10 tasks} & \textbf{20 tasks} & \textbf{5 tasks} &  \textbf{10 tasks} & \textbf{20 tasks} \\
\hline
MUC&38.45&19.57&15.65&18.95&15.47&\phantom{0}9.14\\
+TASS &49.17 \textcolor{red}{(+10.72)}&40.34 \textcolor{red} {(+20.77)}&37.86 \textcolor{red}{(+22.21)}&32.47 \textcolor{red}{(+13.46)}&30.13 \textcolor{red}{(+14.66)}&27.70  \textcolor{red}{(+18.56)}\\ \hline
IL2A&55.13&45.32&45.24&36.77&34.53&28.68\\
+TASS &58.74 \textcolor{red}{(+3.61)}&53.24 \textcolor{red}{(+7.92)}&53.07 \textcolor{red}{(+7.83)}&42.49 \textcolor{red}{(+5.72)}&41.34 \textcolor{red}{(+6.81)}&40.59 \textcolor{red}{(+11.91)}\\ \hline
PASS&55.67&49.03&48.48&41.58 &39.28& 32.78\\
+TASS &59.10 \textcolor{red}{(+3.43)}&54.45 \textcolor{red}{(+5.42)}&52.37 \textcolor{red}{(+3.89)}&44.05 \textcolor{red}{(+2.47)} &43.06 \textcolor{red}{(+3.78)}&42.57 \textcolor{red}{(+9.79)}\\ \hline
SSRE&56.33&55.01&50.47&41.45 &40.07&39.25\\
+TASS &59.26 \textcolor{red}{(+2.93)}&57.93 \textcolor{red}{(+2.92)}&53.78 \textcolor{red}{(+3.31)}&44.13 \textcolor{red}{(+2.68)} &43.86 \textcolor{red}{(+3.79)}&43.55 \textcolor{red}{(+4.30)}\\
\hline 
	\end{tabular} }
	\caption{Performance gain in top-1 accuracy by applying TASS to other EFCIL methods in a plug-and-play way. Absolute gains are indicated in \textcolor{red}{(red)}.}
	\label{tab:plug}
\end{table*}
\begin{table*}[t]
	\centering
	\small
	\renewcommand{\arraystretch}{1.1}
  \setlength\tabcolsep{1.2mm}
	\resizebox{1\textwidth}{!}{
\begin{tabular}{c|c|c|c|c|c|c|c|c|c|c|c|c|c|c|c|c|c|c|c}\hline
\multicolumn{2}{c|}{\textbf{Dataset}}&\multicolumn{9}{c|}{\textbf{CIFAR100} }&\multicolumn{9}{c}{\textbf{TinyImageNet}} \\ \hline
	   \multicolumn{2}{c|}{\textbf{Setting}} & \multicolumn{3}{c|}{\textbf{5 tasks}} &  \multicolumn{3}{c|}{\textbf{10 tasks}} & \multicolumn{3}{c|}{\textbf{20 tasks}} & \multicolumn{3}{c|}{\textbf{5 tasks}} &  \multicolumn{3}{c|}{\textbf{10 tasks}} & \multicolumn{3}{c}{\textbf{20 tasks}} \\
\hline
\multicolumn{2}{c|}{\textbf{Method}}&Avg$\uparrow$&Last$\uparrow$&$F\downarrow$&Avg$\uparrow$&Last$\uparrow$&$F\downarrow$&Avg$\uparrow$&Last$\uparrow$&$F\downarrow$&Avg$\uparrow$&Last$\uparrow$&$F\downarrow$&Avg$\uparrow$&Last$\uparrow$&$F\downarrow$&Avg$\uparrow$&Last$\uparrow$&$F\downarrow$\\
\hline
\multirow{5}{*}{\textbf{\rotatebox{90}{~E=20}}}
&iCaRL-CNN\dag&51.07&40.12&42.13&48.66&39.65&45.69&44.43&35.47&43.54&34.64&22.31&36.89&31.15&21.10&36.70&27.90&20.46&45.12\\
&iCaRL-NCM\dag&58.56&49.74&24.90&54.19&45.13&28.32&50.51&40.68&35.53&45.86&33.45&27.15&43.29&33.75&28.89&38.04&28.89&37.40\\
&LUCIR\dag&63.78&55.06&21.00&62.39&50.14&25.12&59.07&48.78&28.65&49.15&37.09&20.61&48.52&36.80&22.25&42.83&32.55&33.74\\
&EEIL\dag&60.37&52.35&23.36&56.05&47.67&26.65&52.34&41.59&32.40&47.12&34.24&25.56&45.01&34.26&25.91&40.50&30.14&35.04\\ 
&RRR\dag&66.43&57.22&18.05&65.78&55.74&18.59&62.43&51.35&18.40&51.20&42.23&16.67&49.54&40.12&21.64&47.46&35.54&29.10\\
\hline
\multirow{6}{*}{\textbf{\textbf{\rotatebox{90}{~E=0}}}}
&LwF\_MC&45.93&36.17&44.23&27.43&50.47&17.04&20.07&15.88&55.46&29.12&17.12&54.26&23.10&12.33&54.37&17.43&\phantom{0}8.75&63.54\\ 
&EWC&16.04&\phantom{0}9.32&60.17&14.70&\phantom{0}8.47&62.53&14.12&\phantom{0}8.23&63.89&18.80&12.71&67.55&15.77&10.12&70.23&12.39&\phantom{0}8.42&75.54\\
&MUC&49.42&38.45&40.28&30.19&19.57&47.56&21.27&15.65&52.65&32.58&17.98&51.46&26.61&14.54&50.21&21.95&12.70&58.00\\
&IL2A&63.22&55.13&23.78&57.65&45.32&30.41&54.90&45.24&30.84&48.17&36.14&25.43&42.10&35.23&28.32&36.79&28.74&35.46\\
&PASS&63.47&55.67&25.20&61.84&49.03&30.25&58.09&48.48&30.61&49.55&41.58&18.04&47.29&39.28&23.11&42.07&32.78&30.55\\
&SSRE&65.88&56.33&18.37&65.04&55.01&19.48&61.70&50.47&18.37&50.39&41.67&17.25&48.93&39.89&22.50&48.17&39.76&26.74\\

&\textbf{TASS} (Ours)&\textbf{68.75}&\textbf{59.26}&\textbf{16.42}&\textbf{67.42}&\textbf{57.93}&\textbf{17.78}&\textbf{62.76}&\textbf{53.78}&\textbf{17.78}&\textbf{55.12}&\textbf{44.13}&\textbf{15.40}&\textbf{54.21}&\textbf{43.86}&\textbf{18.47}&\textbf{52.79}&\textbf{43.55}&\textbf{22.51} \\\hline
	\end{tabular} 
 }
\caption{Average, last top-1 accuracy, and forgetting on CIFAR-100 with different numbers of tasks. Replay-based methods storing 20 exemplars from each previous class are indicate with $\dag$. The best overall results are in \textbf{bold}. We run all experiments three times and report the mean for all metrics.}
	\label{tab:cmp}
 \vspace{-4mm}
  \end{table*}
  
\section{Experimental Results}
\label{sec:experiments}

In this section we first describe our experimental setup and then compare TASS to state-of-the-art methods on several EFCIL benchmarks. In Section~\ref{subsec:add} we give further analysis of the various components of TASS.


\begin{table}[t]
	\centering
	\small
	\renewcommand{\arraystretch}{1.1}
  \setlength\tabcolsep{1.2mm}
	\resizebox{0.5\textwidth}{!}{
\begin{tabular}{c|c|c|c|c|c|c|c|c|c}\hline
\multicolumn{1}{c|}{\textbf{Dataset}}&\multicolumn{9}{c}{\textbf{ImageNet-Subset} }\\ \hline
	   \multicolumn{1}{c|}{\textbf{Setting}} & \multicolumn{3}{c|}{\textbf{5 tasks}} &  \multicolumn{3}{c|}{\textbf{10 tasks}} & \multicolumn{3}{c}{\textbf{20 tasks}} \\
\hline
\multicolumn{1}{c|}{\textbf{Method}}&Avg$\uparrow$&Last$\uparrow$&$F\downarrow$&Avg$\uparrow$&Last$\uparrow$&$F\downarrow$&Avg$\uparrow$&Last$\uparrow$&$F\downarrow$\\
\hline
LwF\_MC&34.86&24.10&49.36&31.18&20.01&53.04&27.54&17.42&56.07\\ 
MUC&40.65&27.89&47.13&35.07&22.65&52.10&31.44&20.12&53.85\\
PASS&63.12&52.61&22.47&61.80&50.44&23.57&55.23&46.07&26.73\\
SSRE&69.54&58.46&17.22&67.69&57.51&18.60&61.23&50.05&23.22\\

\textbf{TASS} (Ours)&\textbf{74.32}&\textbf{63.14}&\textbf{14.37}&\textbf{72.60}&\textbf{57.93}&\textbf{16.09}&\textbf{68.79}&\textbf{57.60}&\textbf{18.41} \\\hline
	\end{tabular} 
 }
\caption{Average, last top-1 accuracy, and forgetting on ImageNet-Subset for different numbers of tasks. We run all experiments three times and report means for all metrics.}
	\label{tab:subset}
  \end{table}

\subsection{Experimental Setup}
We follow standard experimental protocols for EFCIL on three benchmark datasets. 

\minisection{Datasets.} We perform experiments on CIFAR-100~\cite{krizhevsky2009learning}, Tiny-ImageNet~\cite{le2015tiny}, and ImageNet-Subset~\cite{deng2009imagenet}. For most experiments, we train the model on half of the classes for the first task, and then equally distribute the remaining classes across each of the subsequent tasks. The convention we use is: $F+C \times T$ means that the first task contains $F$ classes, and the next $T$ tasks each contain $C$ classes. This is a common configuration for EFCIL used in both PASS~\cite{zhu2021prototype} and SSRE~\cite{zhu2022self}. 

\minisection{State-of-the-art methods.}
Since we focus on EFCIL, we mainly compare with exemplar-free state-of-the-art approaches: SSRE~\cite{zhu2022self}, PASS~\cite{zhu2021prototype}, IL2A~\cite{zhu2021class},  EWC~\cite{kirkpatrick2017overcoming}, LwF-MC~\cite{rebuffi2017icarl}, and MUC~\cite{liu2020more}. To demonstrate the effectiveness of our method, we also compare its performance with several exemplar-based methods like iCaRL (both nearest-mean and CNN)~\cite{rebuffi2017icarl}, EEIL~\cite{castro2018end}, and LUCIR~\cite{hou2019learning}. 
We also compare with RRR~\cite{ebrahimi2021remembering} integrated with SSRE, which focuses on preserving saliency using exemplar replay.

\minisection{Implementation details and metrics.}
We use ResNet-18~\cite{He2016} as a feature extraction backbone. This is the same base network used in SSRE~\cite{zhu2022self} and PASS~\cite{zhu2021prototype}, two state-of-the-art EFCIL approaches. We use the decoder in~\cite{sfnet} to estimate low-level student saliency maps. All experiments are trained from scratch using Adam for 100 epochs with an initial learning rate 0.001. The learning rate is reduced by a factor 10 at epochs 45 and 90. For exemplar-based approaches, we use herding~\cite{rebuffi2017icarl} to select and store 20 samples per class following common settings~\cite{rebuffi2017icarl,hou2019learning}. We implement RRR~\cite{ebrahimi2021remembering} with SSRE to fairly compare it with TASS. For dilated boundary supervision, we set $d$ of the three mid-level boundary dilation stages to be 5\%, 10\% and 15\% of the image size.

We report three common metrics for class incremental learning: the average and last top-1 accuracy, as well as average forgetting for all classes learned up to task $t$. Denoting by $Acc_i$ the accuracy over all learned classes up to and including task $i$, the average accuracy is defined as $Avg = \frac{\sum_{i=1}^T Acc_i}{T}$, and the last accuracy is $Acc_T$. Letting $a_{m,n}$ denotes the accuracy of task $n$ after learning task $m$, the forgetting measure $f_k^i$ of task $i$ after learning task $k$ is computed as $f_k^i=\max_{t\in1,2,...,k-1}(a_{t,i}-a_{k,i})$. The average forgetting $F_k$ is defined as $F_k=\frac{1}{k-1}\sum_{i=1}^{k-1}f_k^i$. 


\subsection{Comparison with the State-of-the-art}

  \begin{figure*}[t]
    \centering
    \includegraphics[scale=0.24]{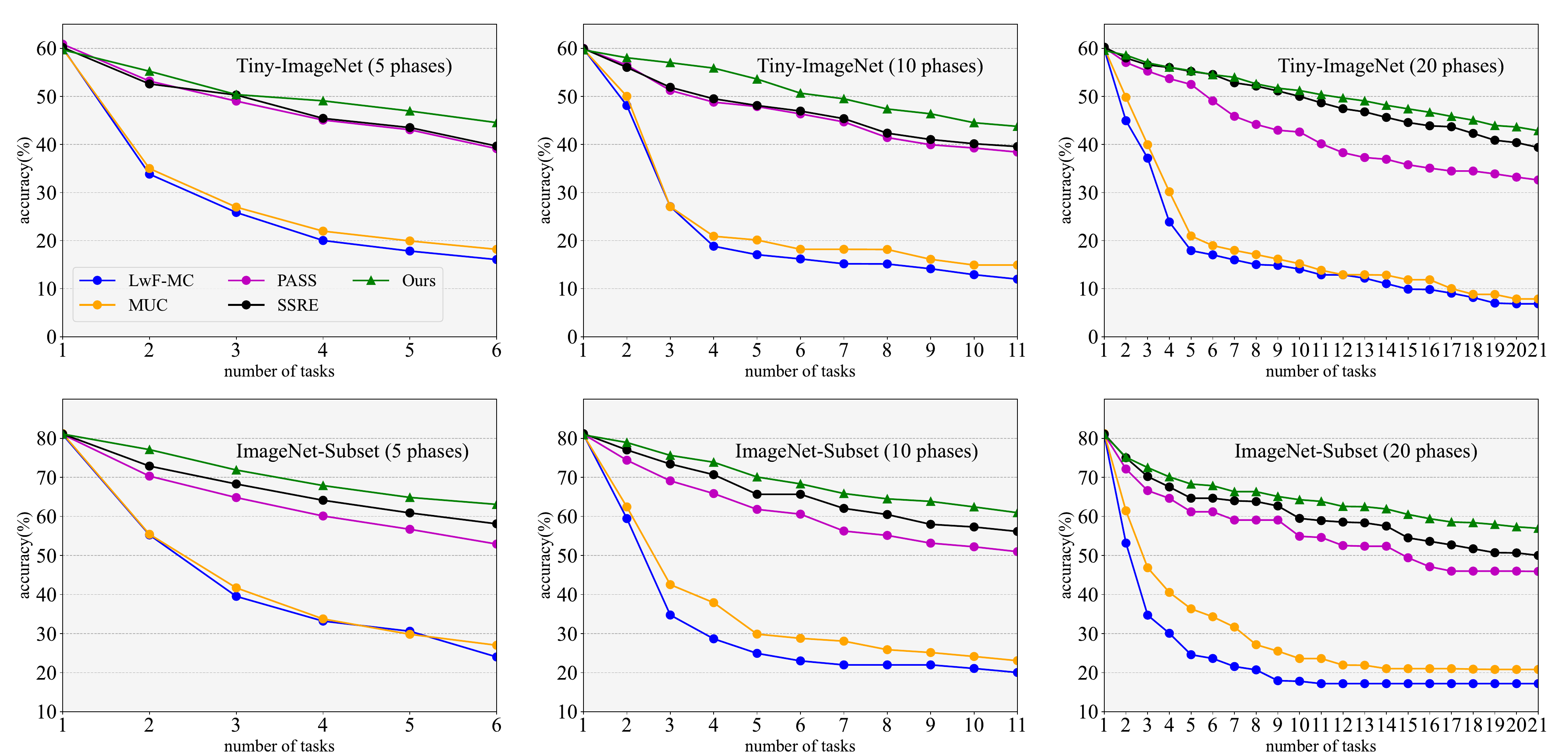}
    \vspace{-8pt}
	\caption{ 
	Results on Tiny-ImageNet and ImageNet-Subset for different numbers of tasks. Our method outperforms others, especially on longer task sequences (i.e. more, but smaller, tasks). 
	}\label{fig:cmp}
\end{figure*}

	

	

We compare TASS with the state-of-the-art on CIFAR-100 in Table~\ref{tab:cmp}. TASS outperforms all exemplar-free approaches. For exemplar-based methods like iCaRL~\cite{rebuffi2017icarl}, EEIL~\cite{castro2018end}, and LUCIR~\cite{hou2019learning}, our method still has significantly better performance. On longer sequences (i.e. 10 and 20 tasks), our method significantly reduces forgetting when learning new classes compared to other EFCIL methods. TASS outperforms the best method SSRE by about 3\% on the last task. This performance improvement can be also observed in terms of average forgetting. 

As we see in  Table~\ref{tab:subset} and Figure~\ref{fig:cmp} for Tiny-ImageNet and ImageNet-Subset, although our method has similar top-1 accuracy on the first task in Figure~\ref{fig:cmp}, it has better performance at most intermediate tasks and also the final one. 
For longer sequences in Figure~\ref{fig:cmp}, the gap between our method and the best baseline is largely consistent, showing the effectiveness of our method at mitigating forgetting. The performance gain in Table~\ref{tab:subset} is larger on Tiny-ImageNet and ImageNet-Subset compared to CIFAR100, and this demonstrates that our method generalizes to datasets with larger images and object scales. It is worth mentioning that TASS also produces results with smaller variance. We believe this to be due to TASS reducing saliency drift to background regions, which may include random noise. 

\minisection{Plug-and-play with other EFCIL methods.} Some existing EFCIL methods, like PASS~\cite{zhu2021prototype}, IL2A~\cite{zhu2021class} and SSRE~\cite{zhu2022self}, focus on reducing forgetting via embedding regularization. Considering the importance of saliency to image classification, it is natural to consider whether TASS can be integrated into these methods. The results in Table~\ref{tab:plug} show the performance gain brought by this integration. Adding TASS doubles the performance for MUC in many cases and significantly improves IL2A and PASS. When we incorporate it into the best baseline SSRE, it yields a consistent gain of about 3\%. These results clearly show that TASS, by explicitly mitigating saliency drift, is complementary to other methods in relieving forgetting. They additionally demonstrate the significance of saliency drift as a cause of catastrophic forgetting in EFCIL.
  
\begin{figure*}[t]
  \centering
  \small
\includegraphics[scale=1.1]{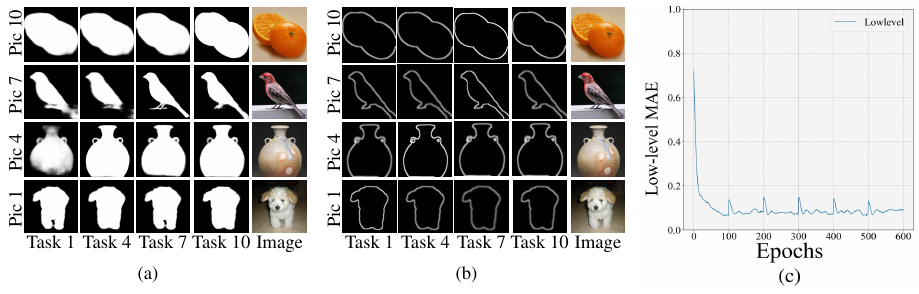}
\caption{Visualization of the saliency (a) and boundary (b) maps from our student encoder-decoder network with original images from different tasks at different stages of incremental learning. Our method produces stable low-level results while reducing forgetting in classification. (c) The MAE loss between the student and teacher network across tasks.
}
		\label{Fig:caf}
  \vspace{-3mm}
	\end{figure*}
	
\newcommand{\addll}[1]{\includegraphics[width=0.11\linewidth]{iccv2023AuthorKit/fig/ourselves/#1.png}}
\newcommand{\addts}[1]{\includegraphics[width=0.46\linewidth]{iccv2023AuthorKit/fig/tsne/#1.png}}

\subsection{Additional Analysis}
\label{subsec:add}
In this section we take a deeper look at the method we propose. If not specified, the results are produced using TASS integrated into SSRE~\cite{zhu2022self}.

\minisection{Ablation Study.} We performed ablations using the 10-task setting on CIFAR-100 (see Table~\ref{tab:abl}). We ablate on both PASS~\cite{zhu2021prototype} and SSRE~\cite{zhu2022self}. Low-level multi-task supervision is crucial and improves by 2.2\% (PASS) and 1.2\% (SSRE). Dilated boundary supervision further boosts performance by about 1-2\%. Saliency noise injection is also helpful for both methods and improves by 1.5\% for PASS. In total, TASS improves baselines by 5.5\% and 2.9\% points, respectively. Note that SSRE is the previous state-of-the-art method and TASS outperforms it by a large margin.


\begin{table}[tp]
	\centering
	\small
	\setlength\tabcolsep{1.3mm}
	\renewcommand{\arraystretch}{1.3}
	
	\begin{tabular}{c|c|c|c|c}
	\hline
	   Method \& Tasks&$\mathcal{L}_{\text{lms}}$&$\mathcal{L}_{\text{dbs}}$&SNI &Accuracy \\ \hline
	   	   Baseline (PASS)&&&&49.0\\
Variants &\checkmark&&&51.2\\
&\checkmark&\checkmark&&53.0\\
&\checkmark&\checkmark&\checkmark&54.5\\
	   \hline
	   Baseline (SSRE)&&&&55.0\\
Variants &\checkmark&&&56.2\\
&\checkmark&\checkmark&&57.3\\
&\checkmark&\checkmark&\checkmark&57.9\\
	   \hline

	\end{tabular}
	\caption{Ablations on each TASS component. Experiments are on CIFAR-100 in the 10-task setting and we report the top-1 accuracy in \% for TASS integrated into PASS and SSRE.  $\mathcal{L}_{\text{dbs}}$ (Eq.~\ref{eq-dbs}), $\mathcal{L}_{\text{lms}}$ (Eq.~\ref{eq-lms}), and SNI denote the three TASS components: Dilated Boundary Supervision, Low-level Multi-task Supervision, and Saliency Noise Injection. 
	}
	\label{tab:abl}
 \vspace{-3mm}
  \end{table}


\minisection{Low-level Multi-task Supervision.} 
To analyze the effect of our proposed low-level saliency supervision, we performed experiments on ImageNet-Subset in the 5 and 10 task settings. We first plot the loss across tasks in Figure~\ref{Fig:caf}(c). 
After learning to predict boundary and saliency maps in the first task, the network maintains good performance for the rest of the 5 task sequence. This shows that the low-level tasks are stable during continual learning. Furthermore, we visualize the results of saliency and boundary map prediction during incremental learning in Figure~\ref{Fig:caf}(a-b). We give some examples of predicted boundary and saliency maps after learning different tasks. Although CIL involves samples of different classes, we see that the low-level outputs are relatively stable and class agnostic. Since the model is able to stably predict these low-level features across tasks, it therefore can preserve useful prior knowledge for continual learning.



\begin{table}[htbp]
    \centering
    \small
    \renewcommand{\arraystretch}{1.1}
    \setlength\tabcolsep{1.2mm}
    \begin{tabular}{c|c|c}
    \hline
    \multicolumn{1}{c|}{\textbf{Metric \& Method}} & Avg$\uparrow$ & Last$\uparrow$ \\
    \hline
    FeTrIL~\cite{petit2023fetril}&65.20&56.34\\    SOPE~\cite{zhu2023self} & 65.84 & 56.80 \\ 
    PRAKA~\cite{shi2023prototype}&68.86&59.20\\ \hline
    TASS (SSRE) & 67.42 & 57.93 \\ 
    TASS (PRAKA)&\textbf{69.70}&\textbf{60.04}\\
    \hline
    \end{tabular} 
    \caption{Average and last accuracy on CIFAR-100 10-task setting.}
    \label{tab:exp}
    \vspace{-0.3cm}
\end{table}

\begin{table}[htbp]
    \centering
    \small
    \renewcommand{\arraystretch}{1.1}
    \setlength\tabcolsep{1.2mm}
    \begin{tabular}{c|c}
    \hline
    \multicolumn{1}{c|}{\textbf{Metric \& Method}} & Avg$\uparrow$ \\
    \hline
    SOPE~\cite{zhu2023self} & 60.20 \\ 
    FeTrIL~\cite{petit2023fetril}&65.00\\   
    \hline
    TASS (FeTrIL)&\textbf{66.03}\\
    \hline
    \end{tabular} 
    \caption{Average accuracy on the ImageNet-Full 10-task setting.}
    \label{tab:imagenet}
    \vspace{-0.3cm}
\end{table}

\minisection{TASS with more methods and benchmarks.} 
Due to the strong generalization capability of our method, we have also applied our TASS paradigm to PRAKA~\cite{shi2023prototype}, resulting in further performance improvements. We give specific experimental comparisons in Table~\ref{tab:exp}. We also compare with FeTrIL~\cite{petit2023fetril} in Table~\ref{tab:exp} and Table~\ref{tab:imagenet} above. Experiments on ImageNet-Full show consistent improvement in Table~\ref{tab:imagenet} above.

\minisection{Quantitative analysis of saliency drift.}  We measure the intersection over union (IoU) between the self-attention map of the last layer in DINO and the Grad-CAM saliency maps for both SSRE and SSRE+TASS. As shown in Table~\ref{tab:iou}, TASS significantly reduces saliency drift in the student model, which again shows the effectiveness of our method for maintaining saliency during CIL.

\begin{table}[htbp]
\centering
\small
\begin{tabular}{c|c|c|c|c}
\hline
\multicolumn{1}{c|}{\textbf{Task}}&1&4&7&10\\ \hline
SSRE&47.4&50.1&56.6&78.5\\
SSRE+TASS&75.2&82.3&88.5&90.1\\
\hline
	\end{tabular} 
\caption{Average IoU (\%) between DINO self-attention maps and student model saliency in the CIFAR-100 10-task.}
	\label{tab:iou}
  \vspace{-0.3cm}
  \end{table}

\section{Conclusions}
\label{sec:conclusions}

In this paper, we propose an approach of task-adaptive saliency guidance for EFCIL. The insight behind TASS is to guide the model to focus on salient regions and inhibit saliency drift. 
We show that robust saliency guidance is crucial to mitigating forgetting across tasks. Experiments demonstrate that TASS is effective and surpasses the state-of-the-art. TASS can be easily combined with other methods, leading to large performance gains over baselines. Qualitative results also show that low-level tasks are stable across different tasks, resulting in less forgetting.

\minisection{Limitations and Future Work.} In this work, we introduce the auxiliary saliency knowledge across tasks for better CIL performance, which may raise the concern of unfair comparison. However, we believe it is important to explore external knowledge to make the CIL system applicable in real applications. Additionally, our method can be easily integrated with other methods and the low-level auxiliary knowledge takes negligible cost and is easy to obtain. We will explore leveraging other forms of knowledge in future work, such as pre-trained visual and large language models. 

\minisection{Acknowledgments}
This work is funded by  
NSFC (NO. 62206135, 62225604), Young Elite Scientists Sponsorship Program by CAST (2023QNRC001), and the Fundamental Research Funds for the Central Universities 
(Nankai Universitiy, 070-63233085). 
Computation is supported by the Supercomputing Center of Nankai University.

{
    \small
    \bibliographystyle{ieeenat_fullname}
    \bibliography{main}
}
\clearpage
\newcommand{\addsn}[1]{\includegraphics[width=0.2\linewidth]{iccv2023AuthorKit/fig/saln/#1.png}}
\begin{figure*}[t]
  \centering
  \small
  \setlength\tabcolsep{0.2mm}
    \renewcommand\arraystretch{0.6}
    \begin{tabular}{ccccccc}
    \addsn{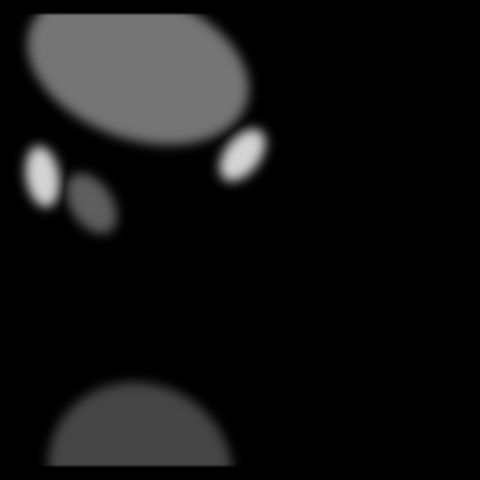}&\addsn{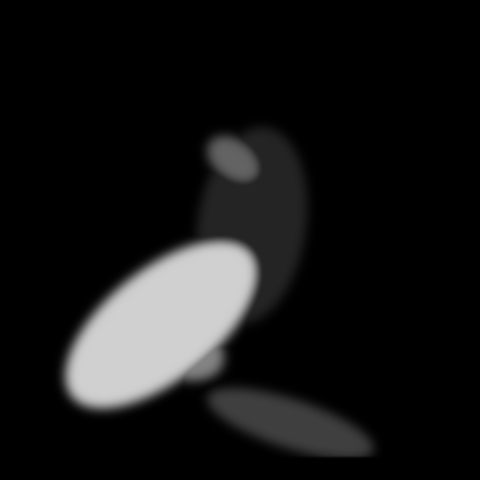}&\addsn{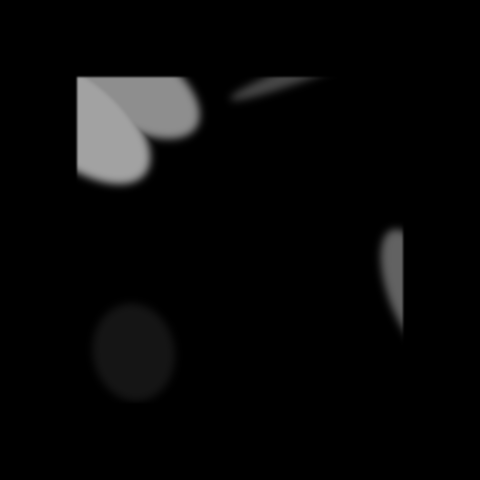}&\addsn{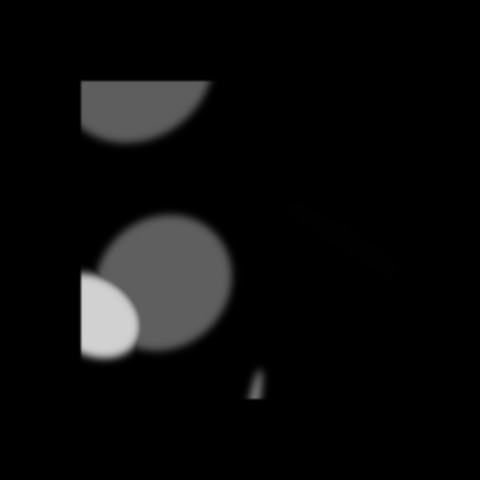}&\addsn{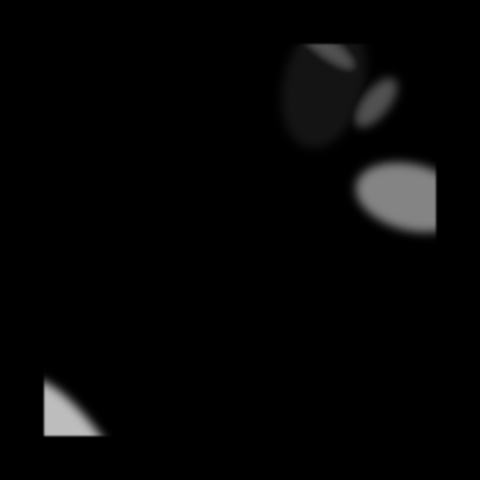}\\
    \end{tabular}

\vspace{-1pt}
		\caption{Visualization of some generated saliency noise maps.
		}
		\label{fig:saln}
	\end{figure*}

\appendix
\section{Further Ablation Studies}
\paragraph{Ablation on saliency methods.}\label{ssm}
To show the generalization of TASS, we use several methods to compute saliency maps and report the results in Table~\ref{tab:stusal}. Grad-CAM performs the best, although other methods yield performance gains, demonstrating the effectiveness of TASS.
\begin{table}[t]
	\centering
	\small
	\setlength\tabcolsep{1.3mm}
	\renewcommand{\arraystretch}{1.3}
	
	\begin{tabular}{c|c|c|c}
	\hline
	 Method  &5 Tasks&10 Tasks&20 Tasks\\ \hline
    baseline (SSRE)&40.2 &40.0&39.3\\
	   CAM &41.2&40.7&40.4  \\
	   SmoothGrad &42.1&41.0&40.4 \\
	   Grad-CAM&44.1&43.9&43.5 \\
	   \hline
	\end{tabular}
	\caption{
	Ablation on methods for generating saliency maps on Tiny-Imagenet.
	}
	\label{tab:stusal}
  \end{table}

\paragraph{Ablation on low-level target maps.}
In the manuscript we use CSNet~\cite{21PAMI-Sal100K} to compute all the pre-trained saliency and boundary maps because it is very lightweight. Compared to our main model, the pre-trained  model has fewer than 1\% parameters and requires 1.5\% of the FLOPs (as shown in Table \ref{tab:flo}). Note that we compute all low-level maps offline before new tasks, and so the extra FLOPs should be amortized over the number of epochs. Therefore, the additional FLOPs required by the low-level model is only about 0.015\% of the main model, which is negligible in practice.

To show the effectiveness of TASS, we perform an ablation on the low-level maps. We replace them with the Grad-CAM generated from a ResNet-152 network. To avoid information leakage, ResNet-152 was trained from scratch. Before each new task, we first train it only on task data and use the Grad-CAM output to supervise saliency in our incremental model. From Table~\ref{tab:nlm} we see that TASS still outperforms other methods. Moreover, TASS is applicable to other models for generating saliency maps, (e.g. DFI~\cite{liu2020dynamic} or PoolNet~\cite{liu2019simple}) with more parameters, and produces even better performance with larger networks.

  \begin{table}[t]
	\centering
	\small
	\setlength\tabcolsep{1.3mm}
	\renewcommand{\arraystretch}{1.3}
	
	\begin{tabular}{c|c|c}
	\hline
	   Model&Parameter(M)&FLOPS(G)\\ \hline
	   Ours &17.9&0.78 \\
	   Pre-trained salient model &0.0941&0.012 \\
	   \hline
	\end{tabular}
	\caption{
	Parameters and FLOPs of the pre-trained salient model. FLOPs are computed using $3\times 32 \times 32$ images.
	}
	\label{tab:flo}
  \end{table}
  
\begin{table}
	\centering
	\small
	\setlength\tabcolsep{1.3mm}
	\renewcommand{\arraystretch}{1.3}
	
	\begin{tabular}{c|c}
	\hline
	   Low-level source (Method)&Accuracy (\%)\\ \hline
	   PASS&39.3\\
	   SSRE&40.0\\
	   ResNet152 (Ours) &42.1 \\
	   CSNet (Ours) &43.9 \\
	   PoolNet (Ours) & 44.2\\
	   DFI (Ours) & 44.4 \\
	   \hline
	\end{tabular}
	\caption{
	Ablation on low-level saliency maps on Tiny-ImageNet with 10 tasks.
	}
	\label{tab:nlm}
  \end{table}
  
\paragraph{Ablation on method architecture and salient model pretraining.}
We select PASS~\cite{zhu2021prototype} as our baseline method to apply TASS to (as shown in Table~\ref{tab:para} and Table~\ref{tab:pretrain}). Experiments in these two tables are conducted on ImageNet-Subset with 5 tasks. Since some methods use ImageNet pretrained weights for better saliency map estimation, we train CSNet~\cite{21PAMI-Sal100K} from scratch on the dataset (with and without pretraining) for salient object detection~\cite{yan2013hierarchical,li2014secrets,yang2013saliency}. This allows us to verify that no information leakage happens due to pretraining the saliency network on ImageNet. The low-level network without pretraining works almost as well as pretraining the saliency network on ImageNet. We also compare the number of parameters of different methods in Table~\ref{tab:para}. This shows that adding network capacity for PASS from ResNet-18 to ResNet-32 with more parameters only improves the performance marginally. Ours with ResNet-18 based on PASS achieves a significant gain surpassing SSRE which has more parameters.

\begin{table}
	\centering
	\small
	\setlength\tabcolsep{1.3mm}
	\renewcommand{\arraystretch}{1.3}
	
	\begin{tabular}{c|c |c}\hline
	   Method&Parameter (M) & Accuracy (\%)   \\ \hline
	   PASS-Res18&14.5&50.4\\
	   PASS-Res32&21.7&51.2\\
	   SSRE-Res18&19.4&58.7\\
	   Ours-Res18&17.9&61.5\\
	   \hline
	\end{tabular}
	\caption{Comparison of different method network architectures. Method-Res18 denotes applying Method with ResNet18 as its backbone.}
	\label{tab:para}
  \end{table}
  
\begin{table}
	\centering
	\small
	\setlength\tabcolsep{1.3mm}
	\renewcommand{\arraystretch}{1.3}
	
	\begin{tabular}{c|c}\hline
	   Method& Accuracy (\%)   \\ \hline
	   No pretraining &61.5\\
	   Pretrained salient detection model&62.0
	   \\\hline
	\end{tabular}
	\caption{Ablation on salient detection network pretraining.}
	\label{tab:pretrain}
  \end{table}

\paragraph{Our approach in non-DFCIL scenarios.}
We apply our saliency supervision in a non-DFCIL scenario using PASS in Table~\ref{tab:ndf} by including 20 exemplars per class. TASS boosts performance significantly here as well.
\begin{table}[h]
	\centering
	\small
	\renewcommand{\arraystretch}{1.1}
  \setlength\tabcolsep{1.2mm}
	
	\vspace{-0.2cm}
	\begin{tabular}{cccccccccc} 
\hline
	  Method&buffer size&Acc (\%)\\ \hline
	  PASS&20&52.36\\
	  PASS+TASS&20&\textbf{55.75}\\
\hline
	\end{tabular} 
  \vspace{-0.2cm}
	\caption{TASS on a non-DFCIL scenario.
	}
 \vspace{-0.5cm}
	\label{tab:ndf}
  \end{table}

\paragraph{Hyper-parameters of multiple losses.}
In Eq. 5 we weight all loss terms equally. As suggested,
we explore more options in Table~\ref{tab:weight}. Tuning further improves the performance slightly, but we stick with $\lambda_{\mathrm{CIL}} =\lambda_{\mathrm{lm}}=\lambda_{\mathrm{dbs}}=1.0$ for convenience. $\sqrt{N}$ in Eq. 2, where $N$ is the number of pixels, is used to normalize the L2 distance. 

\begin{table}[h]
	\centering
	\small
	\renewcommand{\arraystretch}{1.1}
  \setlength\tabcolsep{1.2mm}
	
	\vspace{-0.2cm}
	\begin{tabular}{cccccccccc} 
\hline
	  $\lambda_{\mathrm{CIL}}$&$\lambda_{\mathrm{lm}}$&$\lambda_{\mathrm{dbs}}$&Acc(\%)\\\hline
	  1&1&1&55.01\\
	  0.1&1&1&\textbf{55.27}\\
	  1&0.1&1&54.22\\
	  1&1&0.1&54.31\\
	  \hline
	  
\hline
	\end{tabular} 
  \vspace{-0.2cm}
	\caption{Hyper-parameters of multiple losses for SSRE+TASS.
	}
	\label{tab:weight}
  \end{table}

\begin{table}[h]
	\centering
	\small
	\renewcommand{\arraystretch}{1.1}
  \setlength\tabcolsep{1.2mm}

	\begin{tabular}{cccccccccc} 
\hline
	 Method&&L&D&S&LD&LS&DS&LDS\\ \hline
	 PASS&49.0&51.2&50.6&51.4&53.0&52.6&53.7&54.5\\
      SSRE&55.0&56.2&55.8&56.7&57.3&57.0&57.6& 57.9\\
\hline
	\end{tabular} 
 \vspace{-0.2cm}
	\caption{LDS represent Low-level multi-task supervision, Dilated boundary supervision, and Saliency noise injection, respectively.  
	}
	\label{tab:fa}
  \end{table}

\paragraph{All loss permutations ablation.}
We give all possible combinations of all three loss terms in Table~\ref{tab:fa}. These results show that each component contributes to the final performance and that a combination of them performs best.

\paragraph{Class division in experimental protocol.}
We follow conventional experimental setups from previous works like PASS and SSRE to divide the classes of the dataset as $F+C \times T$ with $F=50$. As suggested, we evaluate different options for $F$ in Table~\ref{tab:split}. TASS shows consistent gain compared to the baseline under all settings.

\begin{table}[t]
	\centering
	\small
	\renewcommand{\arraystretch}{1.1}
  \setlength\tabcolsep{1.2mm}

	\begin{tabular}{cccccccccc} 
\hline
	 F \& Acc (\%)&50&30&10&0\\ \hline
	  PASS&49.03$\pm$0.9&46.78$\pm$0.9&44.65$\pm$1.0&40.27$\pm$1.0\\PASS+TASS&54.45$\pm$0.4&51.22$\pm$0.5&48.58$\pm$0.5&44.30$\pm$0.5\\
	
\hline
	\end{tabular} 
  \vspace{-0.3cm}
	\caption{Ablation on $F$ with $T=10$. 
	}
 \vspace{-0.7cm}
	\label{tab:split}
  \end{table}

\section{More Visualizaions on TASS}
\label{gsn}
\minisection{Saliency Noise.}
For each ellipse there are 6 dimensions: the center coordinate $(x,y)$, the rotation angle $\alpha$, the mask weight $w$, and the major and minor axes $(a,b)$. $x$, $y$, $\alpha$ and $w$ are sampled from a uniform distribution over ranges: $x\in[0,H)$, $y\in[0,W)$, $\alpha\in[0,2\pi)$, $w\in[0,1]$. $H$ and $W$ denote the height and width of input images. To generate ellipses of appropriate size, we draw the major and minor axes from a Gaussian distribution with $\mu_{a} = \max(H,W)/2$, $\sigma_a = \max(H,W)/6$, $\mu_{b} = \min(H,W)/2$, $\sigma_b = \min(H,W)/6$. The sampled $a,b$ is clipped to$[0,max(H,W)/2]$ and $[0,min(H,W)/2]$, respectively. For each ellipse, we create a saliency map $S_i$. We repeat this random generation process 3-5 times and apply an element-wise max operation on the $S_i$ to obtain a single saliency map $S$. Then we crop and resize $S$ to the original image size, with crop size sampled from a uniform distribution in $[\min(H,W)/2, \min(H,W)]$, introducing center-aware saliency noise to the network for training. Finally, we apply a Gaussian blur on $S$ to better simulate a realistic saliency map. The kernel size for Gaussian blurring is the closest odd integer to $\min(H,W)/20$. For each encoder feature map, 10\% of randomly selected channels are directly masked with $S$, where each selected channel will have an independent $S$. We visualize several generated samples in Figure~\ref{fig:saln}.

\begin{figure}
	\centering
	\small
	\renewcommand{\arraystretch}{1.0}
  \setlength\tabcolsep{0.5mm}
	
\begin{tabular}{cccc}
    \rotatebox{90}{~~~~~~~~~~~Ours}&\addts{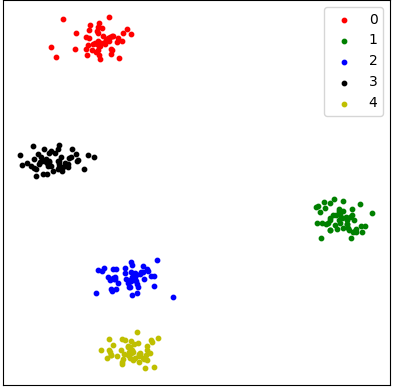}&\addts{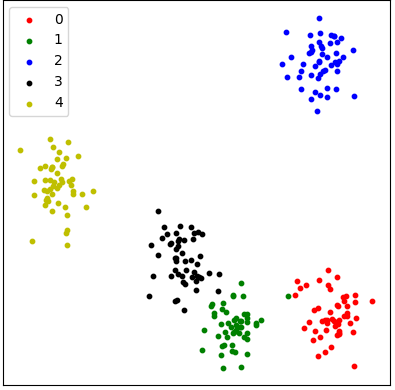}\\
    \rotatebox{90}{~~~~~~Baseline }&\addts{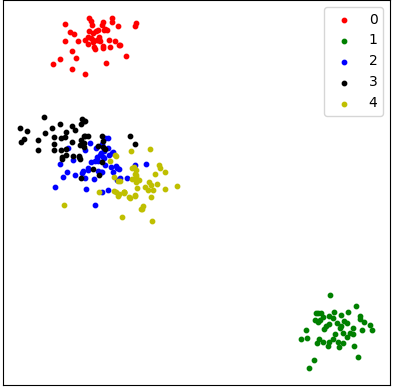}&\addts{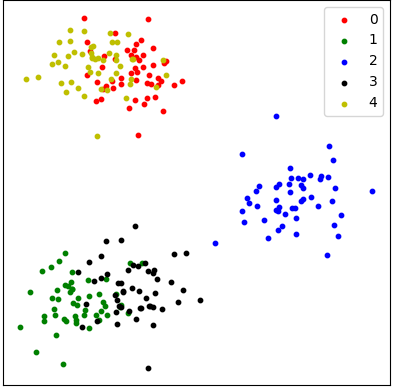}\\
    &Base task&Last task
    \end{tabular}
    \caption{Visualization of the embedding $F_{\phi}(x)$ with and without TASS. Compared to the baseline, our method preserves more discriminative representations.
    }
	\label{Fig:salv}
\end{figure}

\minisection{Embedding Visualization.}
Since our method helps the model focus on the foreground, more class-specific pixels contribute to the embedding. Thus embeddings are more discriminative and contain less distracting background information. In Figure~\ref{Fig:salv} we use t-SNE to visualize embeddings of five initial classes after learning the base and last task in the 10-task setting on ImageNet-Subset. At the base task, both Baseline (SSRE) and Ours (SSRE+TASS) perform well. After the last task, it is clear that TASS helps maintain discriminative features between tasks while the Baseline has overlapping embeddings.
\end{document}